\pdfoutput=1

\documentclass[11pt]{article}

\usepackage{emnlp2023}

\usepackage{times}
\usepackage{latexsym}
\usepackage{graphicx}
\usepackage{colortbl}
\usepackage{multirow}
\usepackage{booktabs}

\usepackage{algorithm}
\usepackage{algpseudocode}
\usepackage{float}
\usepackage{hyperref}
\usepackage[nolist]{acronym} 
\usepackage{subcaption}

\newcommand\blfootnote[1]{%
  \begingroup
  \renewcommand\thefootnote{}\footnote{#1}%
  \addtocounter{footnote}{-1}%
  \endgroup
}

\usepackage{comment}
\usepackage[T1]{fontenc}

\usepackage[utf8]{inputenc}

\usepackage{microtype}

\usepackage{inconsolata}

\title{A tailored Handwritten-Text-Recognition System for Medieval Latin}

\author{Philipp Koch$^1$\textsuperscript{$\diamondsuit$} 
        \And Gilary Vera Nuñez$^1$\textsuperscript{$\diamondsuit$}
        \And Esteban Garces Arias$^1$\textsuperscript{$\spadesuit$} 
        \AND Christian Heumann$^1$\textsuperscript{$\spadesuit$}
        \And Matthias Schöffel$^2$\textsuperscript{$\clubsuit$}
        \And Alexander Häberlin$^{2,3}$\textsuperscript{$\heartsuit$}
        \AND Matthias Aßenmacher$^{1,4}$\textsuperscript{$\spadesuit$} \\ \\
        $^1$ Department of Statistics, LMU, Munich, Germany \\ 
        $^2$ Bavarian Academy of Sciences, BAdW, Munich, Germany \\
        $^3$ Universität Zürich, Zurich, Switzerland\\ 
        $^4$ Munich Center for Machine Learning (MCML), LMU, Munich, Germany\\
        \small \textsuperscript{$\diamondsuit$}\texttt{\{philipp.koch,gi.vera\}@campus.lmu.de} \; \textsuperscript{$\clubsuit$}\texttt{matthias.schoeffel@badw.de} \; \textsuperscript{$\heartsuit$}\texttt{alexander.haeberlin@sglp.uzh.ch}\\
        \small \textsuperscript{$\spadesuit$}\texttt{\{esteban.garcesarias,chris,matthias\}@stat.uni-muenchen.de}
        }
\begin{document}
\maketitle


\begin{abstract} 
The Bavarian Academy of Sciences and Humanities aims to digitize its Medieval Latin Dictionary. This dictionary entails record cards referring to lemmas in medieval Latin, a low-resource language. A crucial step of the digitization process is the \ac{HTR} of the handwritten lemmas found on these record cards. In our work, we introduce an end-to-end pipeline, tailored to the medieval Latin dictionary, for locating, extracting, and transcribing the lemmas. We employ two \ac{SOTA} image segmentation models to prepare the initial data set for the HTR task. Furthermore, we experiment with different transformer-based models and conduct a set of experiments to explore the capabilities of different combinations of vision encoders with a GPT-2 decoder. Additionally, we also apply extensive data augmentation resulting in a highly competitive model. The best-performing setup achieved a \ac{CER} of 0.015, which is even superior to the commercial Google Cloud Vision model, and shows more stable performance.
\end{abstract}


\section{Introduction}

\blfootnote{This paper has been accepted at the \href{https://www.ancientnlp.com/alp2023/}{First Workshop on Ancient Language Processing}, co-located with RANLP 2023.}

\begin{figure}
    \centering
    \includegraphics[width=.5\textwidth]{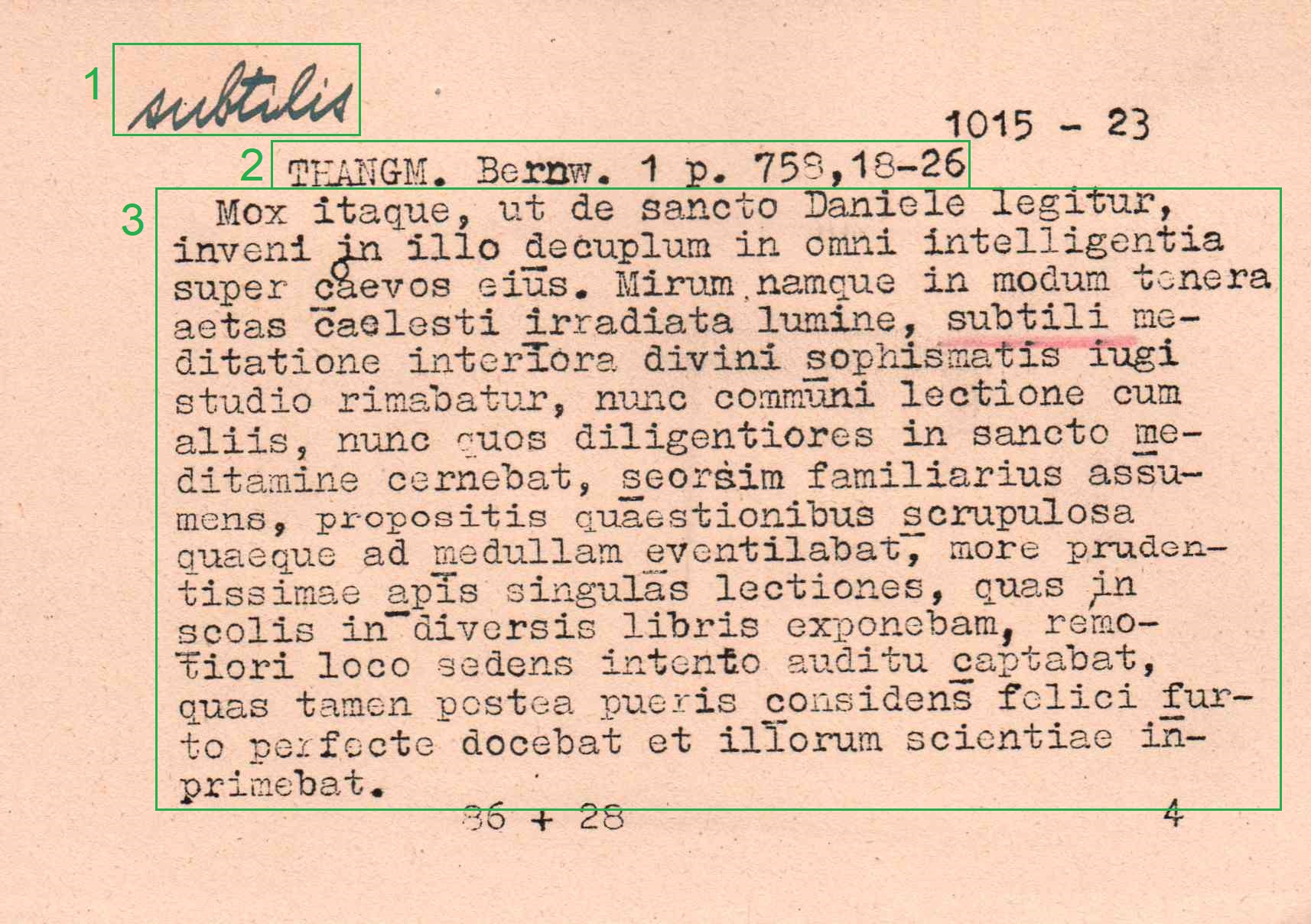}
    \caption{Record card from the MLW data set.}
    \label{fig:sample-rec-card}
\end{figure}

The \ac{MLW}\footnote{In German: \textit{\textbf{M}ittel\textbf{l}ateinisches \textbf{W}örterbuch (MLW)}}, located at the Bavarian Academy of Sciences, deals with Latin texts that were created between 500 and 1280 in the German-speaking region. The foundations for this project have been developed from 1948 onwards and since then, the dictionary has been continuously published in individual partial editions since 1959. Currently, the letter \textit{S} is being worked on in particular. The basis of the dictionary consists of 50 selected texts that have been fully transcribed onto DIN-A6 sheets (record cards) constituting about 40\% of the note material. Later, another 2,500 texts were excerpted and transcribed manually onto DIN-A6 record cards, using a typewriter (cf. Fig. \ref{fig:sample-rec-card}). In addition, there are so-called "index cards", a type of record card, that helps to uncover often hundreds of additional references. In total, it is estimated that 1.3 million reference points have been recorded for the \ac{MLW}. These record cards were sorted alphabetically by the first letter of the keyword (lemma), and serve as the foundation for creating a dictionary. By 2025, at least half of the note material is planned to be scanned and recorded in a database.

To digitize the material, the lemmas - always found in the upper left corner of the record cards, either hand- or machine-written - must be extracted from the cards and recognized using an \ac{OCR} or \ac{HTR} procedure. Around 200,000 record cards have been scanned (cf. Fig. \ref{fig:sample-rec-card}), and annotated with their respective lemma.
The accurate extraction and transcription of the lemma present a challenge, which is further compounded by the limited resources available for medieval Latin. To address this, we develop an end-to-end pipeline that begins by extracting the lemma from the record cards and subsequently utilizes an elaborated \ac{HTR} system to recognize the text.

\paragraph{Contributions} 

\begin{enumerate}
\item We present a novel end-to-end \ac{HTR} pipeline specifically designed for detecting and transcribing handwritten medieval Latin text. Notably, it surpasses commercial applications currently considered \ac{SOTA} for related tasks.
\item We successfully train a detection model without relying on human-annotated bounding boxes for the lemmas.
\item We conduct extensive experiments to compare various vision encoders and evaluate the effectiveness of data augmentation techniques.
\item We make our codebase, models, and data sets publicly available.
\end{enumerate}


\section{Related Work}
\label{sec:related}

We provide an overview of the field of \ac{HTR}, which is the main challenge of this work. We also deal with an instance of object detection to prepare the training data. However, since this problem is only an intermediate step and not the aim of this work, we do not cover it extensively. We refer to the survey of \citet{zaidi2021survey} for a detailed overview.

The recognition of handwritten text differs from \ac{OCR} insofar as it needs to deal with less standardized data.
Previous approaches have focused on applying deep learning to tackle these tasks. Here, the objective of \ac{CTC} \citep{gravesctc2006} comes into play. \ac{CTC} is a technique in which a neural network -- initially a \ac{RNN} but other networks might also be used \citep{chaudhary2022easter20} -- is trained to predict a matrix of conditional transition probabilities. The input image, represented as a vector representation through a \ac{CNN}, is fed to the network, and for each input (i.e. the activation maps of the CNN) the network predicts the character. After obtaining the probabilities, a matrix of conditional transition probabilities can be constructed. A unique void character is introduced to avoid false repetitions, and the final sequence can be obtained and compared with the ground truth. Since many sequences can be obtained from the matrix, the network is trained to maximize the correct conditional transition probabilities. During inference time, the model cannot compute the path of all likely sequences but instead needs to predict the class just in time. For this purpose, search algorithms like beam search or infix search are used.

\ac{CTC}, combined with CNNs and RNNs, often yielded competitive results, such as shown by \citet{PuigcerverCRNN2017} and \citet{BlucheCRNN2017}. Furthermore, approaches applying only CNNs and \ac{CTC} also exist \citep{Chaudhary2021EASTER,chaudhary2022easter20}. The model Easter2.0 achieved competitive results on the IAM data set \citep{MartiIAM2002}, a data set consisting of English handwritten text and being widely used for \ac{HTR}.

A recent work that achieved \ac{SOTA} results on the IAM data set is the TrOCR model \citep{li2022trocr}, based on the transformer \citep{vaswani2017attention}. The model consists of a vision encoder and a text decoder, deviating from previous approaches in which CNNs and RNNs have been primarily used. The input is processed through the encoder and represented in vector space. A language model for decoding subsequently produces the text to be predicted. However, with the emergence of the transformer in the vision domain \citep{dosovitskiy2021image,bao2022BEiT}, end-to-end modeling has become possible. 
In the work of \citet{barrere2022}, another transformer-based model is applied for HTR. The main difference to TrOCR is a different embedding technique for visual features based on a \ac{CNN}. Furthermore, the model also applies \ac{CTC} during training. The results have also been shown to be competitive on the IAM data set. 
\citet{diaz2021rethinking} compared different encoder-decoder models' performance on HTR. In their study, they used different models in the encoder and decoder parts, so a transformer encoder is used before using a \ac{CTC}-based decoder. Furthermore, they found that a transformer encoder and a \ac{CTC}-trained decoder enriched with a language model achieved \ac{SOTA} results on the IAM data set.

The TrOCR framework has been successfully applied to historical data akin to our task. In the work of \citet{ströbel2022transformerbased}, a TrOCR instance was fine-tuned to handwritten Latin from the 16th century \citep[][referred to as \textit{Gwalther}]{peter_stotz_2021_4780947}, achieving competitive results.


\section{Data}
\label{sec:data}


Our data set comprises 114,653 images (18,9 GB), corresponding to 3,507 distinct lemmas. All images are in RGB, but not uniform in size, i.e. height, and width differ from image to image. Additionally, the information on the corresponding lemma (i.e., the ground truth) is available for each image as well as the dictionary's vocabulary.

\paragraph{Image data}
Figure \ref{fig:sample-rec-card} shows one (arbitrarily chosen) sample from the data set. Most record cards follow the same structure being composed of three main parts, highlighted via green boxes. The first one (1), and the one we deem most challenging, is the lemma, which is always located in the upper left corner of the record card. The second part (2) is the index of the text where the lemma is found. The third part (3) contains a text extract in which the word (corresponding to the lemma) occurs in context.

\paragraph{Lemma Annotation}

\begin{figure}[hb]
    \centering
    \includegraphics[width=.5\textwidth]{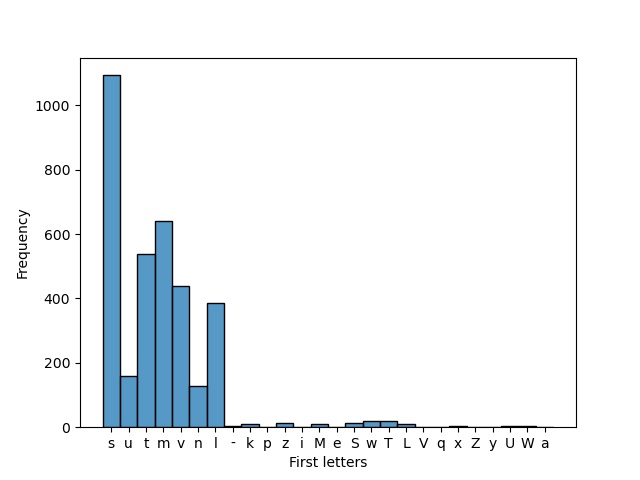}
    \caption{Distribution of the first letters of the lemmas.}
    \label{fig:freqs}
\end{figure}

\begin{figure}[hb]
    \centering
    \includegraphics[width=.5\textwidth]{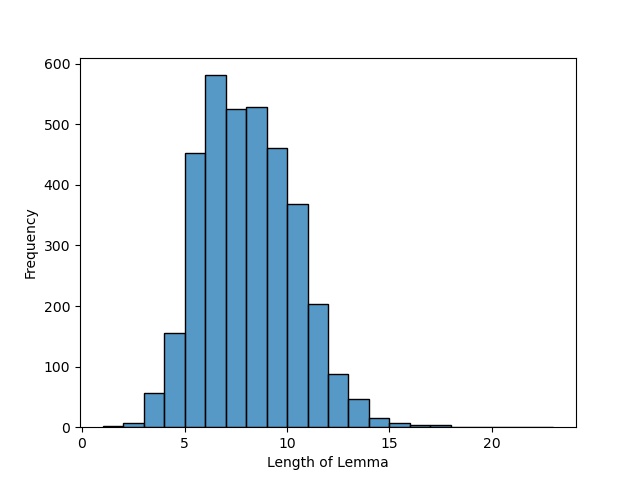}
    \caption{Length distribution of the lemmas.}
    \label{fig:freqs2}
\end{figure}

Our analysis is based on lemma annotations on an image level, i.e. which lemma is on the corresponding record cards. There is a total of 17 different first letters, eight of which are each upper- and lowercase, as well as one special character. The capitalization of a word plays a crucial role since a word’s meaning changes depending on capitalization. Since the majority of our data stems from the \textit{S}-series of the dictionary, most lemmas start with the letter \texttt{"s"}. Likewise, we found a large number of lemmas starting with the letters \texttt{"m"}, \texttt{"v"}, \texttt{"t"}, \texttt{"u"}, \texttt{"l"}, and \texttt{"n"} (cf. Fig. \ref{fig:freqs}).
We also analyzed the number of record cards available per lemma. In this analysis, we found that some lemmas are under-represented in the data set, while a few constitute a large chunk of the data.

A total of 2,420 lemmas (69\%) were found to have ten record cards or less; 854 lemmas (24.4\%), between 10 and 100 record cards, and just 233 lemmas (6.6\%), more than 100 record cards. It is worth mentioning that 1,123 lemmas (approx. 36.7\%) had only one record card.

Finally, we analyze the length of the lemmas (cf. Fig. \ref{fig:freqs2}). We observe lemmas from a length of one character up to a maximum of 19 characters. The average length of the lemmas lies between five and six characters. The presence of such long lemmas motivated the decision of additionally using a weighted metric for model evaluation, as will be explained in Section \ref{sec:metrics}.


\section{Lemma Extraction Pipeline}
\label{sec:pipe}

In this section, we delve into the details of the custom-designed pipeline for the extraction of the lemma from the record cards.  

\begin{figure}[H]
    \centering
    \includegraphics[width=.5\textwidth]{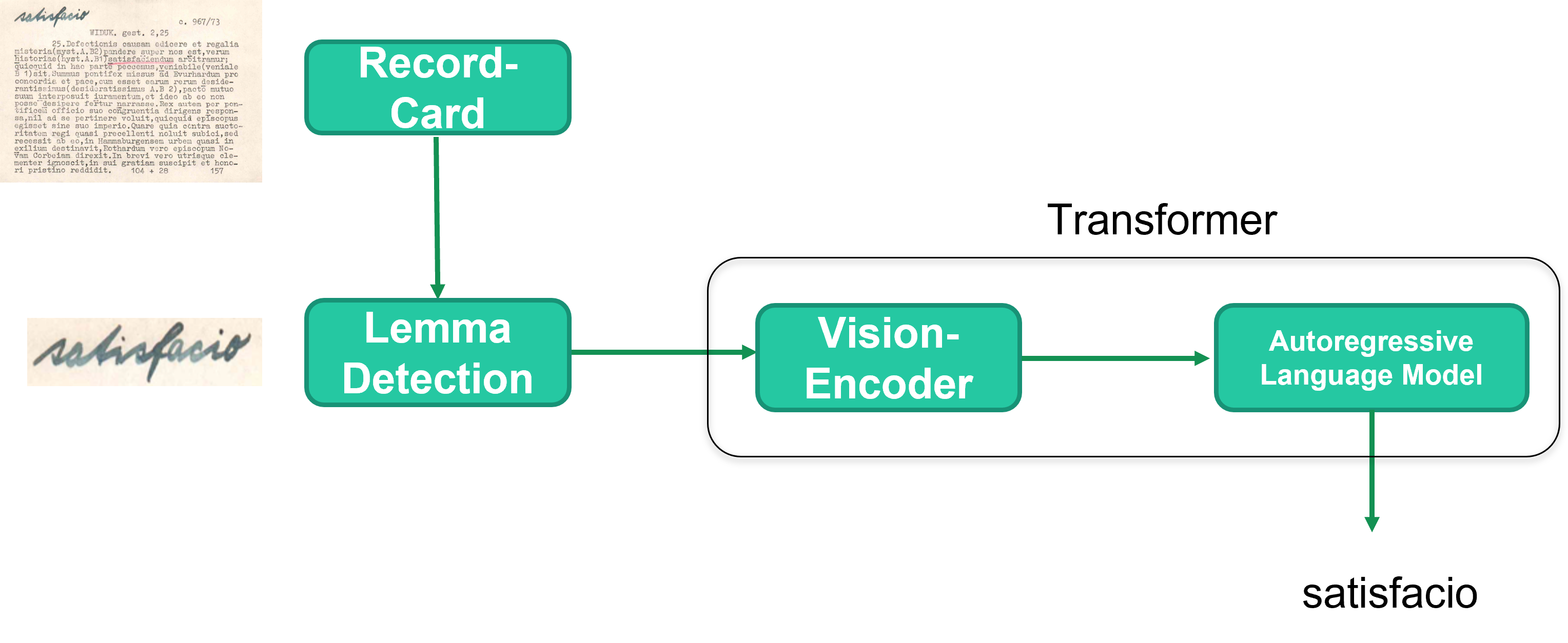}
    \caption{Visualization of the designed pipeline, encompassing three building blocks: (1) the visual detection of the lemma, followed by (2) the encoding in the latent space, and (3) the decoding into plain text.}
    \label{fig:pipe}
\end{figure}

\subsection{Visual Detection}
\label{sec:localiie}

Due to the data structure, we are confronted with the problem of finding suitable bounding boxes to extract the lemmas from the upper left of the record cards. When using the entire record cards for the recognition task, the majority of the image is noise, making model training significantly more difficult. Since the lemmas are not annotated with their exact locations, training a custom object detection model for extraction is not feasible. In order to still retrieve the locations of the bounding boxes for some lemmas, we transform the problem into an instance of visual grounding by providing a model with an image and the description of an object in the image, upon which it is expected to return the object’s location. We use the \ac{OFA} transformer \citep{wang2022ofa}, fine-tuned on RefCOCO \citep{kazemzadeh-etal-2014-referitgame}. To ensure the quality of the extracted lemma, we experiment with multiple prompts and examine their results (cf. Appendix \ref{a:bbox}).

After obtaining a training data set of 20,000 instances, each of them annotated with bounding boxes, we train a YOLOv8 model \citep{Jocher_YOLO_by_Ultralytics_2023} based on the \ac{YOLO} architecture \citep{redmon2016look}. The model predictions from our \ac{YOLO} model, are then subject to two post-processing steps (described in the following) to ensure the quality of the images. 

\paragraph{Multiple Bounding Boxes:} For 17,674 images (15.42\% of the data), the model predicted more than just one bounding box. We visually examined the cases and found that other handwritten text was often recognized as a lemma, sometimes scattered throughout the record cards (e.g. upper or lower right). The distribution of the bounding boxes throughout the record cards is displayed in Figure \ref{fig:yolo_all_bb} (Appendix \ref{a:yolo_train_inf}).

\paragraph{Missing Bounding Box:} We visually examined the 202 cases where no bounding box was detected, some stemming from machine writing (instead of cursive handwriting) or scanning errors. For some images that follow the standard layout of the record cards, the model also failed. We disregard this set constituting less than 0.2\% of the entire data set. 

\paragraph{Determining the Bounding Box}
Taking all aspects into account, we introduce two rules to determine the appropriate bounding box: (1) choose the largest bounding box, and (2) the bounding box has to be in the upper left quarter of the entire image. The result after applying these rules is displayed in Figure \ref{fig:yolo_all_bb_after} (Appendix \ref{a:yolo_train_inf}). The final data set consists of 114,451 samples, exhibiting a difference of the 202 samples to the initial 114,653 image-label pairs. We make our data available on HuggingFace.\footnote{\url{https://huggingface.co/misoda}}

\subsection{HTR Model}
\label{sec:htr}

We use a transformer as the main model akin to TrOCR. For the encoder, we consider three different architectures, while we use GPT-2 \citep{radford2019language} as a decoder model for all setups. All models are trained from scratch, although we use pre-trained image processors for the encoder models and train a tokenizer for our custom alphabet.

\paragraph{Tokenizer}
We use a customized byte-level BPE \citep{sennrich2016neural} tokenizer for the dictionary's vocabulary. The tokenizer is trained on the labels from our data set. 

\paragraph{Vision Encoders}

We consider three different encoder architectures, namely \ac{ViT} \citep{dosovitskiy2021image}, \ac{BEiT} \citep{bao2022BEiT}, and \ac{Swin} \citep{liu2021swin}. 

\ac{ViT} is a transformer-encoder-based model employing 16 x 16 image patching for transforming images into sequences. Additionally, a class patch is concatenated to the sequence of patches, which is used for classification tasks and entails general information about the sequence, similar to the \texttt{CLS} token in BERT \cite{devlin-etal-2019-bert}. For training, a feed-forward neural network is stacked on top of the encoder, serving as an adapter between the encoder and the targets during pre-training. 

Akin to ViT, \ac{BEiT} builds on image patching and an image vocabulary. For pre-training, masked image modeling is introduced, inspired by masked language modeling from natural language processing \citep{devlin-etal-2019-bert}. Further, the encoder exhibits a visual vocabulary, and a Variational Auto Encoder \citep{kingma2022autoencoding} is trained in advance to encode an image to a lower dimension. 
The decoder reproduces the image from the latent codes which can be used as a visual vocabulary, and based on this vocabulary, an image can be represented through a sequence of visual tokens. For our purpose, we train the model from scratch and only re-use the pre-trained image processor.


A problem in the vision domain is the high dimensionality and the often spatially related information in the data. Splitting the image into large patches might break the often fine-grained relations of entities on the images. To overcome this issue, self-attention can be applied more fine-grained, however, this results in higher computational cost. In \ac{Swin}, this issue is tackled using a new encoder block structure, differing substantially from the other transformers. To overcome the above-mentioned problems, the self-attention mechanism is applied differently to account for different aspects of the image. In the lower layers, the image is divided into small patches, and the self-attention mechanism is applied to the small patches in windows. These windows are shifted in upper layers to connect the different patches. Furthermore, the windows are enlarged in upper layers producing a hierarchical representation.
Our model uses a newly initialized Swin transformer alongside a pre-trained image processor.

\paragraph{Text Decoder}
GPT-2 \citep{radford2019language} is a decoder-only transformer that has shown competitive capabilities in text generation. A decoder can only be trained to predict the next token based on the previous sequence while relying on encoded information from the encoder. Since the problem of predicting the next token is a classification task, the Cross-Entropy Loss is used. GPT-2 has been shown to be able to capture the underlying patterns and structures of natural language, making it capable of generating coherent and contextually appropriate text. Due to the overall strong performance of GPT-2, we chose to use it as a decoder for our model. We train it from scratch, i.e., we do not use the pre-trained weights since we deal with a specific task in a low-resource language setting.

\paragraph{Implementation Details}
We use the HuggingFace transformers library \citep{wolf2020huggingfaces} and PyTorch \citep{paszke2019pytorch} to train the \ac{HTR} pipeline. Our codebase, containing all scripts (experiments and training) is available via GitHub\footnote{\url{https://github.com/slds-lmu/mlw-htr}}, and the final model is on pypi.\footnote{\url{https://pypi.org/project/mlw-lectiomat/}}
All the experiments were conducted using a Tesla V100 GPU (16 GB).


\section{Experiments}
\label{sec:exp}


\subsection{Standard Training}
\label{sec:standard}

After shuffling the data, we randomly split it into a train (85\% -- 97,283 samples) and a test (15\% -- 17,168 samples) set. In the train split, 94.53\% (3,315) of the lemmas are present. For all training procedures, we use the AdamW optimizer \citep{loshchilov2019decoupled} and did not engage in hyperparameter tuning. Further details are reported in Appendix \ref{a:train_details}. For standard training, the model is trained using a data set that includes the cut images from the record cards as input and their respective lemmas as the labels to be predicted. We train each of the models for a total of 5 epochs.


\subsection{Data Augmentation}
\label{sec:augment}

Augmentation is a common technique used in deep learning to diversify the training data by applying different modifications without changing the underlying semantics of the data. The goal of augmentation is to provide the model with a diverse set of examples, helping it generalize better and improve performance. \citet{yang2022image} show that augmentation can notably improve the results of deep learning models. We apply on-the-fly augmentation to our data due to the large data set size.

To provide maximal modification and increase the diversity of the training data, different augmentation techniques are applied at random on-the-fly. These techniques include random rotation, blurring, or modifications related to color perception. Since the augmentation is applied on-the-fly, it is necessary to increase the number of epochs so that the model has also enough opportunities to observe the original, unmodified data and the augmented variations. We increased the number of epochs to 20 (compared to 5 for the standard training).

\begin{figure}[ht]
    \centering
    \includegraphics[width=0.4\textwidth]{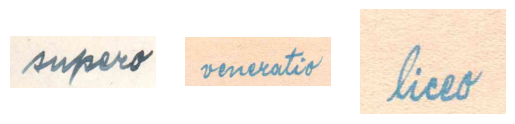}
    \caption{Original lemmas without any modification.}
    \label{fig:aug_orig}
\end{figure}

We use three different augmentation pipelines, one of which is randomly chosen with $p = \frac13$. In the following, we will illustrate each of them at the example of the lemmas shown in Figure \ref{fig:aug_orig}.

\begin{figure}[ht]
    \centering
    \includegraphics[width=0.4\textwidth]{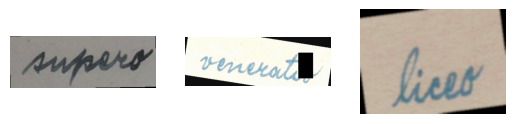}
    \caption{Exemplary samples from pipeline A.}
    \label{fig:aug_1}
\end{figure}

\paragraph{Pipeline A} For the first pipeline, blurring and modifications to sharpness are applied to the data. The intensity of these modifications is determined randomly and can range from no modification to higher intensity (cf. Fig. \ref{fig:aug_1}). 

\begin{figure}[ht]
    \centering
    \includegraphics[width=0.4\textwidth]{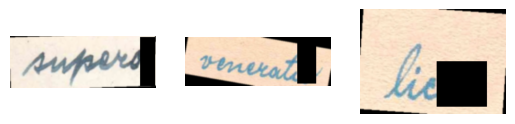}
    \caption{Exemplary samples from pipeline B.}
    \label{fig:aug_2}
\end{figure}

\paragraph{Pipeline B} For the second pipeline, various modifications are applied to alter brightness, contrast, saturation, sharpness, and hue. The specific alterations for each instance are again determined randomly, also including the possibility of no modifications at all (cf. Fig. \ref{fig:aug_2}). 

\begin{figure}[ht]
    \centering
    \includegraphics[width=0.4\textwidth]{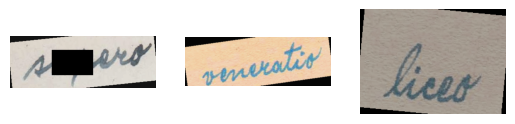}
    \caption{Exemplary samples from pipeline C.}
    \label{fig:aug_0}
\end{figure}

\paragraph{Pipeline C} The third pipeline combines the modifications from the previous two (cf. Fig. \ref{fig:aug_0}).

In addition to the described techniques, all augmentation pipelines include random masking, where rectangles of the images are blackened, and random rotation within a range of -10 to 10 degrees.

\paragraph{Decoder Pre-Training}
We experiment with pre-training the decoder in order to incorporate prior knowledge about the vocabulary we want to predict in the medieval Latin language. After pre-training the decoder on a corpus of the concatenated lemmas, we combine it with the encoder and continue training as described in Section \ref{sec:standard}. While pre-training is performed for a total of 10 epochs, the training of the entire transformer is conducted for 20 epochs. In this approach, the same augmentation techniques, as outlined before, are applied to the training data.


\subsection{Performance metrics}
\label{sec:metrics}

We assess the model performance using the \ac{CER}, which is computed by summing up edit operations and dividing by the length of the lemma.

\begin{equation}
    CER = \frac{S+D+I}{N} = \frac{S+D+I}{S+D+C}, \\
    \label{eq:cer}
\end{equation}

\noindent where \textit{S} is the number of substitutions, \textit{D} is the number of deletions, \textit{I} is number of insertions, \textit{C} is the number of correct characters, and \textit{N} is number of characters in the label. To account for the varying length of the lemmas, we further utilize the weighted \ac{CER}.

\begin{equation}
    Weighted CER = \frac{\sum_{i=1}^n l_i*CER_i}{\sum_{i=1}^n l_i}, 
    \label{eq:cer-weighted}
\end{equation}

\noindent where $l_i=$ is the number of characters of label $i$, and $CER_i$ is the CER for example $i, i=1, \ldots, n$. 


\subsection{Experimental Results}
\label{sec:res}

The main results of our work are reported in Table \ref{tab:results}. The BEiT+GPT-2 architecture achieved the best results in case of the standard training regime, exhibiting a CER of 0.258, followed by Swin+GPT-2 (0.349) and ViT+GPT-2 (0.418). 

Applying the augmentation pipelines, as described in Section \ref{sec:augment}, notably improves model performance compared to the standard training for all three models. The best model with augmentation is Swin+GPT-2, achieving a CER of 0.017. As for the other two models, the CER is 0.073 for ViT+GPT-2 and 0.110 for BEiT+GPT-2.

\begin{table}[h!]
    \centering
    \resizebox{.5\textwidth}{!}{
    \begin{tabular}{lccc}
        \toprule
                                & ViT           & Swin                  & BEiT \\
        \midrule
        Standard                & 0.418         & 0.349                 & 0.258  \\
        + Data Augmentation     & 0.073         & \textbf{0.017}        & 0.110  \\
        + Decoder Pre-Training  & 0.049         & 0.018                 & 0.114  \\
        \bottomrule
    \end{tabular}
    }
    \caption{CER-Results for different encoder configurations.}
    \label{tab:results}
\end{table}

Pre-training of the decoder does, on average, not lead to further improvement. ViT+GPT-2 is the exception, for which the CER drops to 0.049. We observe no improvements for the other models (BEiT+GPT-2: 0.114, Swin+GPT-2: 0.018).

To summarize, the best results are achieved when using a Swin+GPT-2 model with data augmentations, reaching a CER value of 0.017.


\subsection{Ablation Study}
\label{sec:ablation}

To investigate the impact of the data augmentation, we perform three ablations, removing individual steps from the augmentation pipelines. Our ablations include applying modifications to the image regarding sharpness, brightness, color, and blurring (cf. Sec. \ref{sec:augment}). We also apply random rotation and random erasing of some image parts, resulting in black rectangles (masking). To investigate the individual effects of each augmentation technique, we train the model without a specific augmentation method and report the resulting CER. 

\begin{table}[h!]
    \centering
    \resizebox{.5\textwidth}{!}{
    \begin{tabular}{lc}
        \toprule
        Swin+GPT-2 (Full augmentation pipelines) & 0.017             \\
        \midrule
        w/o masking augmentation                & \textbf{0.015}    \\
        w/o rotation augmentation               &           0.021   \\
        w/o color augmentation                  &           0.017   \\
        \bottomrule
    \end{tabular}
    }
    \caption{CER-Results of different model configurations.}
    \label{tab:results_abl}
\end{table}

The results of the ablation study can be seen in Table \ref{tab:results_abl}. 
Excluding the masking step from the pipeline leads to an actual improvement of model performance, such that the CER improves to 0.015.
However, excluding random rotations of the images leads to an increase in CER to 0.021, while augmentation without applying the color-related augmentations results in a CER of 0.017, equal to the initial model trained with all augmentation techniques.
Please note, that only the specific technique was left out at a time, while the other modifications were still in use. From the results of these ablations, we can conclude that adding rotation seems to be a major contribution to prediction quality. Random masking decreases the performance, while color does not seem to have an impact on the performance.


\subsection{Google Cloud Vision Comparison}
\label{sec:gcv}

\begin{figure*}[ht]
    \centering
    \includegraphics[width=\textwidth]{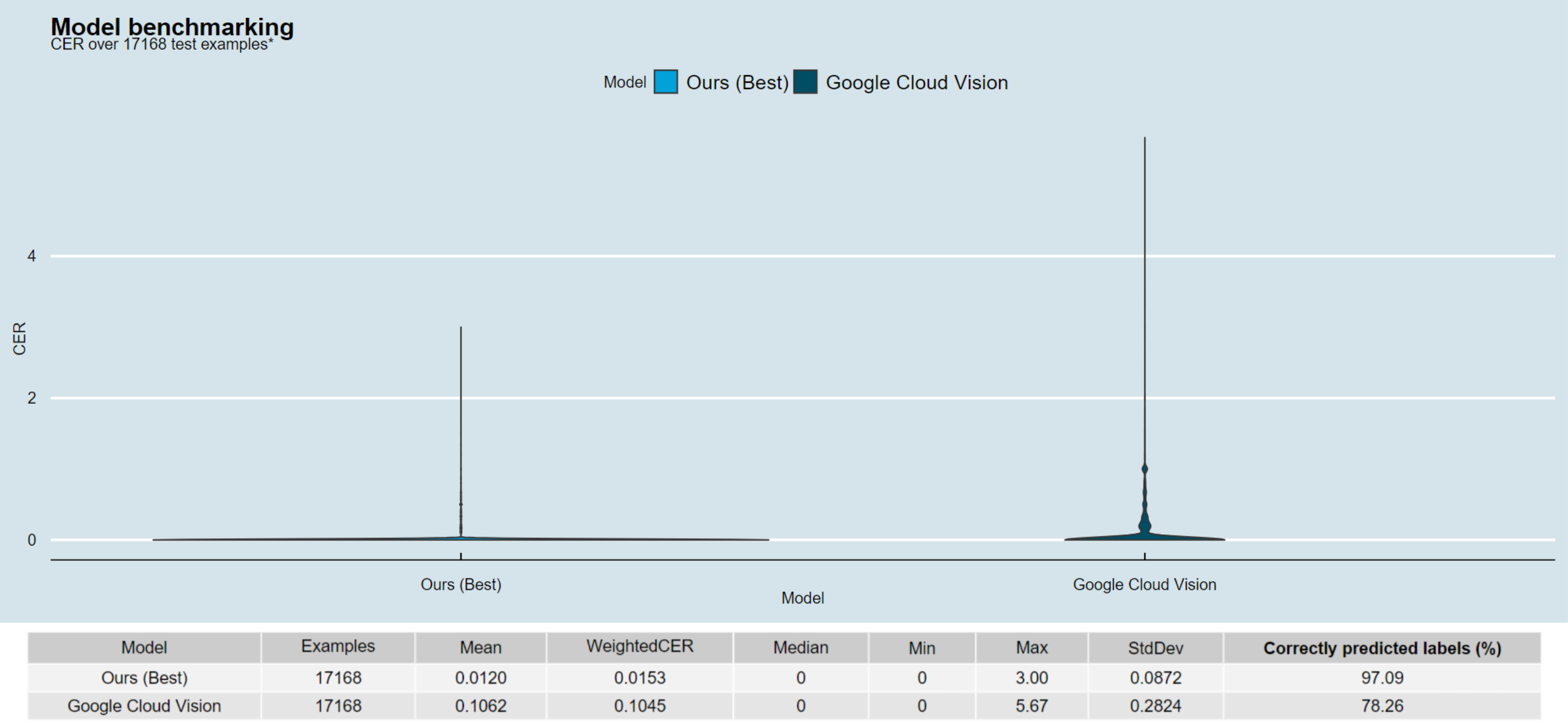}
    \caption{Violin plots for the comparison of our Swin+GPT-2 model (left) to Google Cloud Vision (right).}
    \label{fig:comp_gcv_vs_ours}
\end{figure*}

To compare the results of our model, we decided to use a highly competitive model for HTR, the \href{https://cloud.google.com/vision?hl=de}{\ac{GCV}} model. It is capable of recognizing handwritten text 
and has been proven performative in practical applications \citep{thai_kfz}. As already mentioned, some of the cut record cards which contain the lemmas in our data set contain extra characters and/or suffixes that are not part of the true lemma. We observe that GCV often predicts these extra characters as well. Considering this issue, we decided to post-process the predictions by \ac{GCV} for a fair comparison. This work consisted of deleting extra characters and words after the first word or after a \texttt{'-'} or a \texttt{'('}. Nevertheless, it was not possible to remove all artifacts from the predictions. In some cases, separating them was impossible since the characters were predicted altogether. Figure \ref{fig:comp_gcv_vs_ours} shows the comparison of our model with GCV. 

The violin plots of the (unweighted) CERs show a concentration of the CER values around 0 (= correct prediction) for both models. For our model, the most extreme values are at a CER of 3, for GCV the maximum is nearly twice as high and we observe an overall higher standard deviation compared to our model. Note, that these extreme values originate from the problem that the models sometimes predict too many characters, which are not part of the true (annotated) lemma. To conclude, our best model exhibits a weighted CER of 0.0153, while GCV only reaches 0.1045. Overall, our model correctly predicts 97,09\% of all lemmas, while GCV only does so for 78.26\%.

\subsection{Performance of other \ac{HTR} systems}

Table \ref{tab:res_comp} illustrates the CERs of other systems on different HTR data sets. \citet{ströbel2022transformerbased} use the Rudolph Gwalther data set, while all other papers evaluate their systems on the IAM data set. Our model achieves the lowest CER. However, it must be considered that we did not evaluate the same data set, which makes a direct comparison impossible. In contrast to the other transformer-based models, our best model uses Swin as an encoder which we have not found in other work.

\begin{table}[ht]
    \centering
    \resizebox{.5\textwidth}{!}{
    \begin{tabular}{l|c|c|c}
        \toprule
        Model & CER & Data set & Architecture \\ \midrule
        Ours (Best) & \textbf{0.0153} & MLW  & Transformer \\ \midrule
        TrOCR Large \citep{ströbel2022transformerbased} & 0.0255 & Gwalther & Transformer \\
        TrOCR Large \citep{li2022trocr} & 0.0289 & IAM & Transformer \\
        EASTER2.0 \citep{chaudhary2022easter20} & 0.0621 & IAM & CNN+\ac{CTC} \\
        Light Transformer \citep{barrere2022} & 0.0570 & IAM & CNN+Transformer\\
        Self-Att.+\ac{CTC}+LM \citep{diaz2021rethinking} & 0.0275 & IAM & Trf.+\ac{CTC}+ LM\\
        \bottomrule
    \end{tabular}
    }
    \caption{Performance of contemporary HTR systems evaluated on different data sets.}
    \label{tab:res_comp}
\end{table}


\section{Discussion and Outlook}
\label{sec:outlook}

Due to the focus on recognizing the lemma, we did not experiment with other object detection or image segmentation techniques. Since the record cards include much more information than the one we extracted, we recommend further research into various extraction techniques. With the recent publication of Segment Anything Model, \citet{kirillov2023segment} introduce a model that might be able to extract features from the record cards with much higher accuracy. The next objective could be to extract the inflected lemmas (cf. Sec. \ref{sec:data}).

We did neither experiment with the initial TrOCR architecture nor did we fine-tune a pre-trained TrOCR instance for this task. However, the results of \citet{ströbel2022transformerbased} suggest a strong performance of TrOCR. Thus we also recommend training it on the \ac{MLW} data set. On the other hand, the results of using the Swin encoder indicate a powerful performance compared to the other models we have used. Thus, we also suggest more research into investigating the usage of Swin as an encoder for this task.


\section{Conclusion}
\label{sec:concl}

We present a novel end-to-end pipeline for the Medieval Latin dictionary. Our library includes an image-detection-based model for lemma extraction and a tailored \ac{HTR} model. We experiment with training different configurations of transformers using the ViT, BEiT, and Swin encoders while using a GPT-2 decoder. Employing data augmentation, our best model (Swin+GPT-2) achieves a CER of 0.015. The evaluation of the results exhibits a weaker performance on longer lemmas and on lemmas that appear less frequently in the training data. Further experiments with generative models to produce synthetic data (not reported in the paper) were not successful, however, we recommend further research into the direction of creating synthetic data. To conclude, our approach presents a promising \ac{HTR} solution for Medieval Latin. Future research can build upon our work, and explore its generalizability to other languages and data sets by making use of our pip-installable Python package: \url{https://pypi.org/project/mlw-lectiomat/}


\clearpage

\section*{Limitations}

Our approach has several limitations that can be addressed to improve its efficiency further. There are issues regarding the data set (cf. Sec. \ref{sec:data}) that might be reflected in the model's performance. As discussed in Section \ref{sec:data}, some lemmas are stroked out partially or entirely, introducing a notable noise to the data. Further, handwritten comments or other annotations have been added to some of the record cards, and some images are not correctly labeled, which might have distorted the recognition capabilities of our model.

Since our pipeline was mostly trained on data from the \textit{S}-series of the dictionary, many words starting with other letters were not seen by the model during training. Therefore, the performance of the proposed approach, when applied to other series, remains somewhat uncertain. As elaborated in section \ref{sec:concl}, the model tends to perform weaker on unseen lemmas. Further, there are indications that the model might perform worse on longer lemmas.

The lemma-detection model (YOLOv8) is not guaranteed to predict the correct bounding box for the lemma consistently. 
Errors at this early stage of the pipeline may severely impact the result. Although the failure rate for the training dataset in which no bounding box was predicted is close to zero, the problem can still appear during inference.



\section*{Ethics Statement}

We affirm that our research adheres to the \href{https://www.aclweb.org/portal/content/acl-code-ethics}{ACL Ethics Policy}. This work involves the use of publicly available data sets and does not involve human subjects or any personally identifiable information. We declare that we have no conflicts of interest that could potentially influence the outcomes, interpretations, or conclusions of this research. All funding sources supporting this study are acknowledged. We have made our best effort to document our methodology, experiments, and results accurately and are committed to sharing our code, data, and other relevant resources to foster reproducibility and further advancements in research.

\section*{Acknowledgements}

We wish to thank the Bavarian Academy of Sciences for providing us with the guidance and required access to the handwritten material. This work has been partially funded by the Deutsche Forschungsgemeinschaft (DFG, German Research Foundation) as part of BERD@NFDI - grant number 460037581.

\bibliography{consulting}
\bibliographystyle{acl_natbib}


\clearpage


\appendix

\section*{Appendix}

\section{Annotatong the Bounding Boxes}
\label{a:bbox}

This Appendix holds the details of the Visual Detection part of the pipeline, described in Section \ref{sec:localiie}, and the challenges we were confronted with.

\subsection{The Task}

To annotate the bounding boxes, the model is provided with a prompt describing the lemma and the image. The model then returns a bounding box for the requested object, which is the lemma in our case. Different prompts are described in Table \ref{tab:prompts}.

\begin{table}[ht]
    \centering
    \resizebox{.5\textwidth}{!}{
    \begin{tabular}{c|ll}
        Prompt 1 & & \texttt{Cursive text upper left} \\ \hline
        Prompt 2 & & \texttt{Handwritten cursive word upper left} \\ \hline
        \multirow{2}{*}{Prompt 3}       & Length: 1-5: & \texttt{Blue drawing in the upper left} \\
                                        & Other: &  \texttt{Handwritten cursive word upper left} \\ \hline
        \multirow{2}{*}{Prompt 4}       & Length: 1-6: & \texttt{Blue drawing in the upper left} \\ 
                                        & Other: & \texttt{Handwritten cursive word upper left} \\ \hline
    \end{tabular}
    }
    \caption{Different prompts used for OFA.}
    \label{tab:prompts}
\end{table}

\subsection{Assumption about Bounding Boxes}

Since we do not have any ground truth about the bounding boxes, we rely on heuristics to verify the correctness of the boxes. One such heuristic is the assumed linear relationship between the lemma length and the bounding box's width. While the height of the boxes is assumed to be similar across instances, the lemma length must significantly impact the bounding box's width. To verify the results of the annotation process, we use box plots to visualize the relationship between lemma length and width (cf. Fig. \ref{fig:bp-length-v1} -- \ref{fig:bp-length-v4}).

\subsection{Initial Implementation and Results}

We use the RefCOCO-OFA model\footnote{\href{https://huggingface.co/OFA-Sys/ofa-base-refcoco-fairseq-version}{Huggingface: \texttt{OFA-Base-RefCOCO}}} and modify it four our purposes. Prompt one (cf. Tab. \ref{tab:prompts}) is used to obtain the lemmas for all images.

\begin{figure}[H]
\centering
\begin{subfigure}{0.4\textwidth}
    \includegraphics[width=0.9\linewidth]{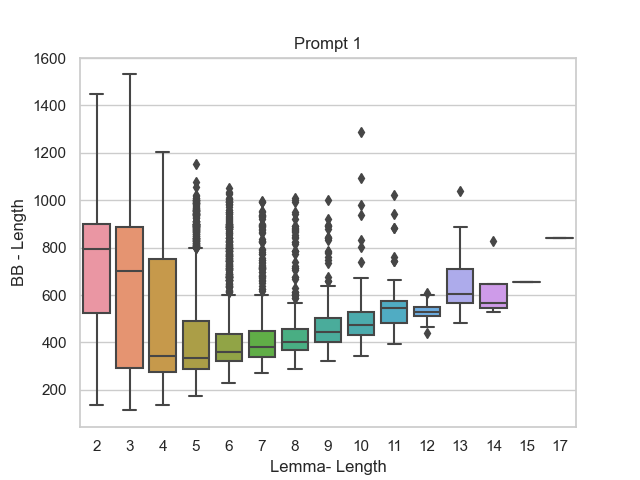}
    \caption{First Prompt}
    \label{fig:bp-length-v1}
\end{subfigure}
\begin{subfigure}{0.4\textwidth}
    \includegraphics[width=0.9\linewidth]{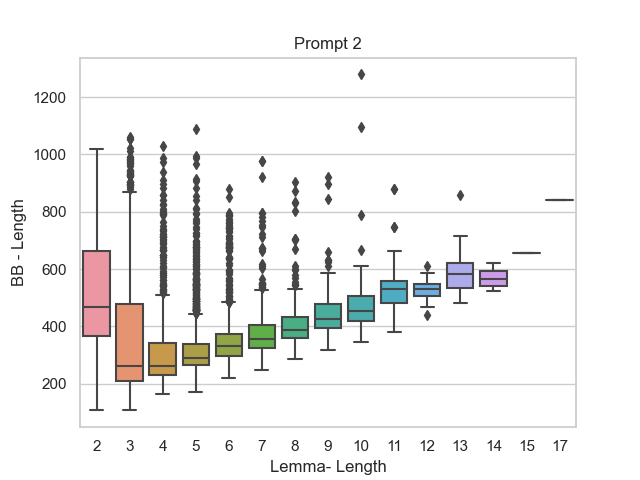} 
    \caption{Second Prompt}
    \label{fig:bp-length-v2}
\end{subfigure}

\begin{subfigure}{0.4\textwidth}
    \includegraphics[width=0.9\linewidth]{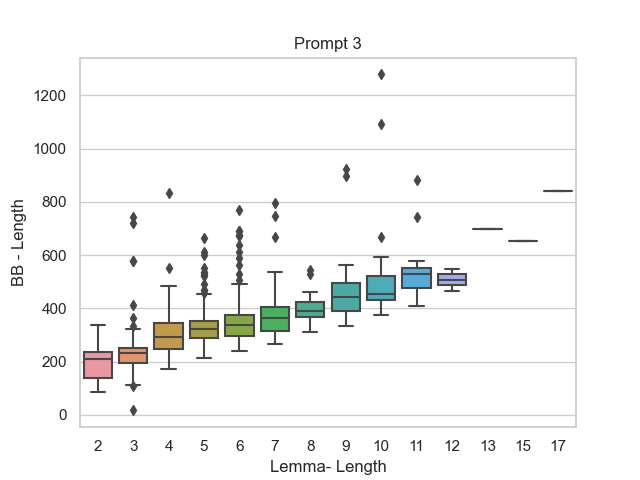}
    \caption{Third Prompt}
    \label{fig:bp-length-v3}
\end{subfigure}
\begin{subfigure}{0.4\textwidth}
    \includegraphics[width=0.9\linewidth]{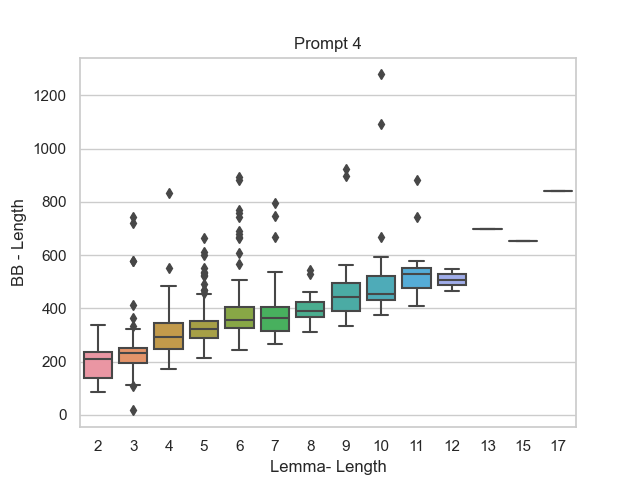} 
    \caption{Fourth and final Prompt}
    \label{fig:bp-length-v4}
\end{subfigure}

\caption{Box-Plots for the width of the bounding boxes based on the lemma's length.}
\label{fig:bp-length}
\end{figure}

After running the model on the first instances with \textit{Prompt 1}, we find that the relationship between the box's width and the lemma length does not look as expected. Figure \ref{fig:bp-length} illustrates this problem. Investigating the short lemmas, we observe that the model often fails to annotate the record cards appropriately. Often other textual objects are annotated, or the bounding box stretches throughout the entire record card. 

\subsection{Two Different Prompts for Shorter and Longer Lemmata}

After different experiments, \textit{Prompt 2} turned out to work appropriately for shorter lemmas, but was, however,  not suitable for longer ones. To combine the strength of both prompts, we apply a conditional prompt based on the length of the lemma using different cut-offs (5 or 6 characters). We find that using \textit{Prompt 4} is the best-suited approach. The analysis of the relationship between the bounding box widths and the length of the lemma for different prompts can be seen in Figure \ref{fig:bp-length}.

\section{YOLO: Training and Inference}
\label{a:yolo_train_inf}

\subsection{Training Results}

\begin{figure}[ht]
    \centering
    \includegraphics[width=.5\textwidth]{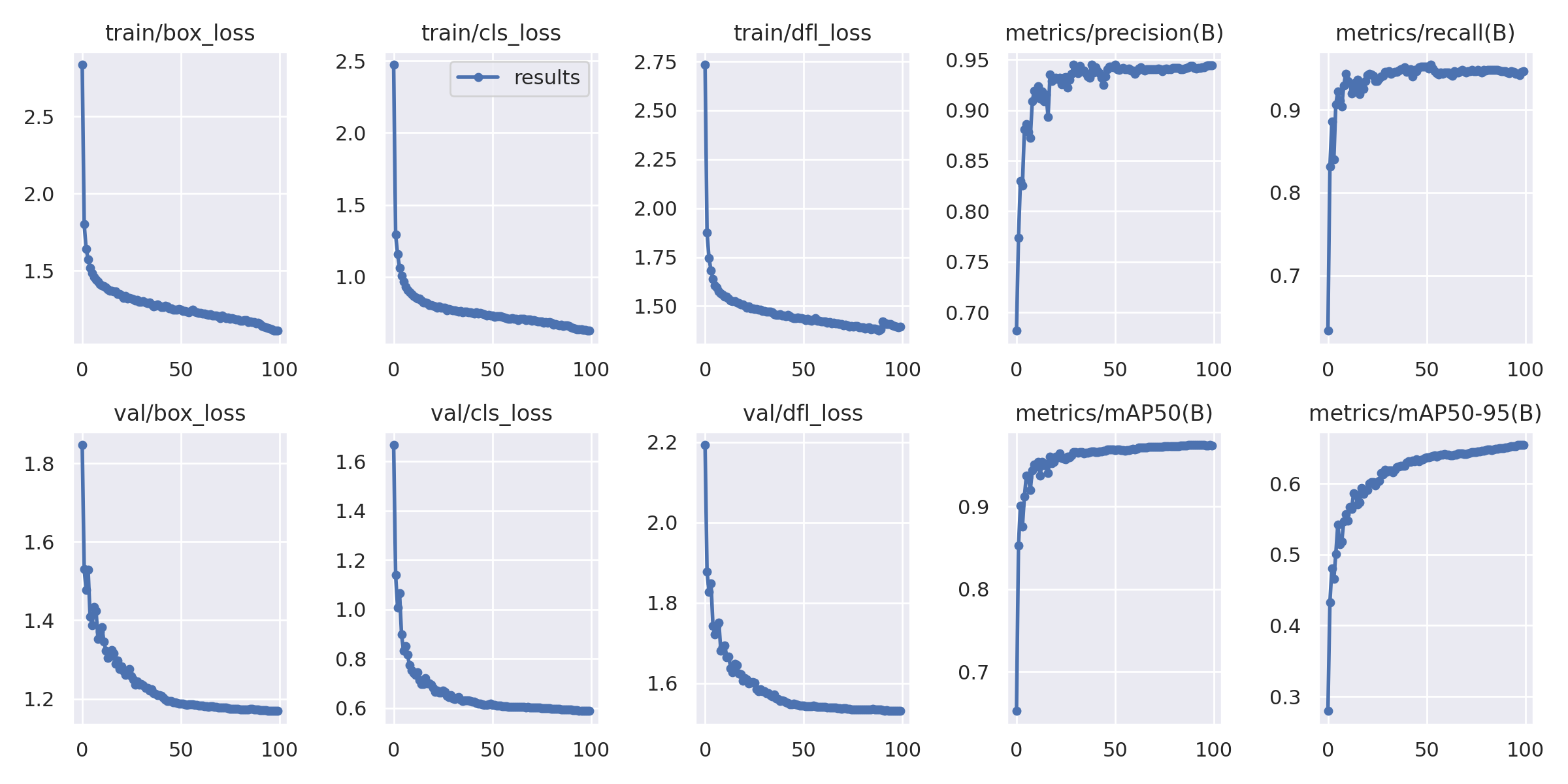}
    \caption{YOLO Training Results.}
    \label{fig:yolo_train_res}
\end{figure}

\subsection{Multiple Lemmas Detected by YOLO}

\begin{figure}[ht]
    \centering
    \includegraphics[width=.5\textwidth]{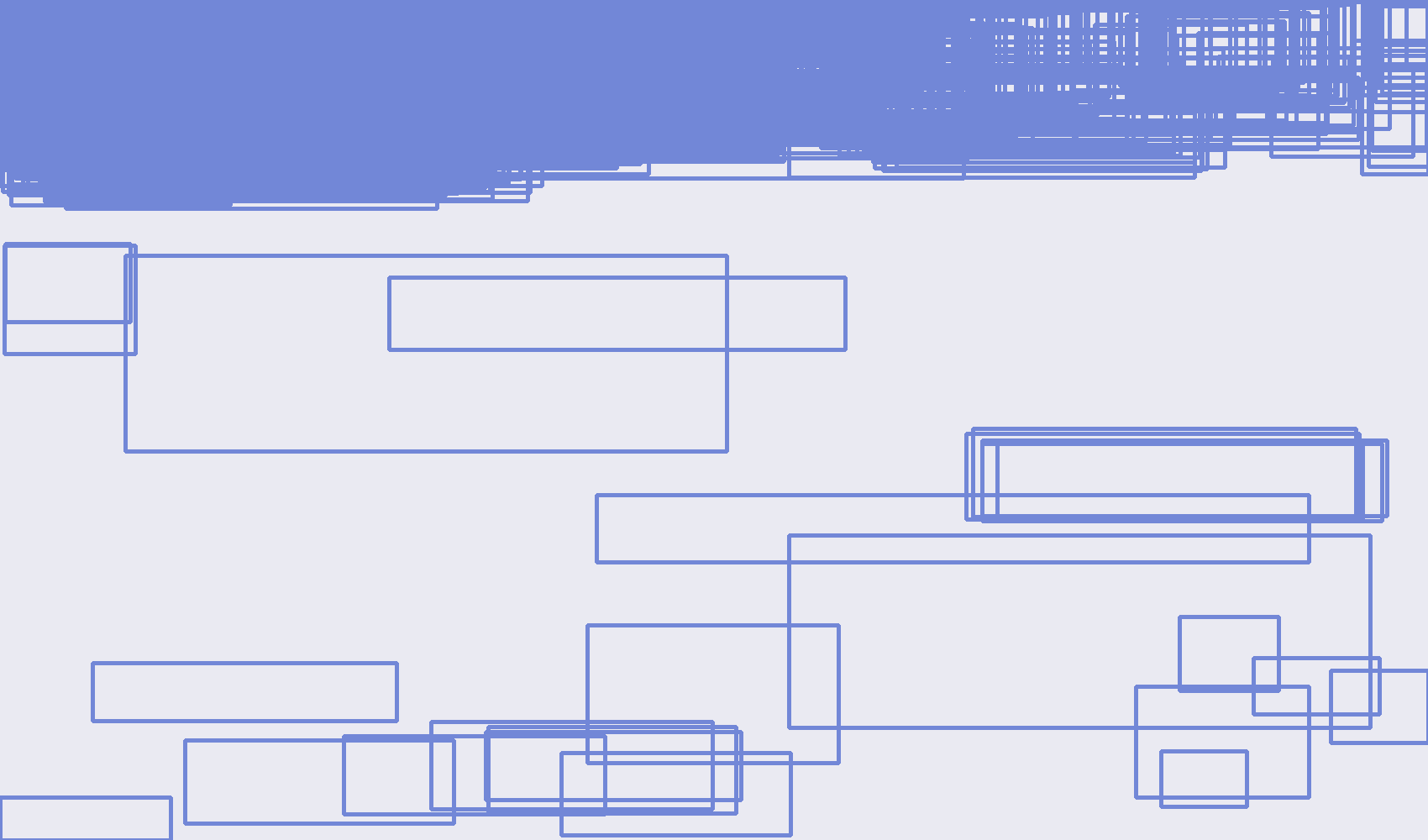}
    \caption{All bounding boxes from instances where YOLO has detected more than one bounding box.}
    \label{fig:yolo_all_bb}
\end{figure}

\begin{figure}[ht]
    \centering
    \includegraphics[width=.5\textwidth]{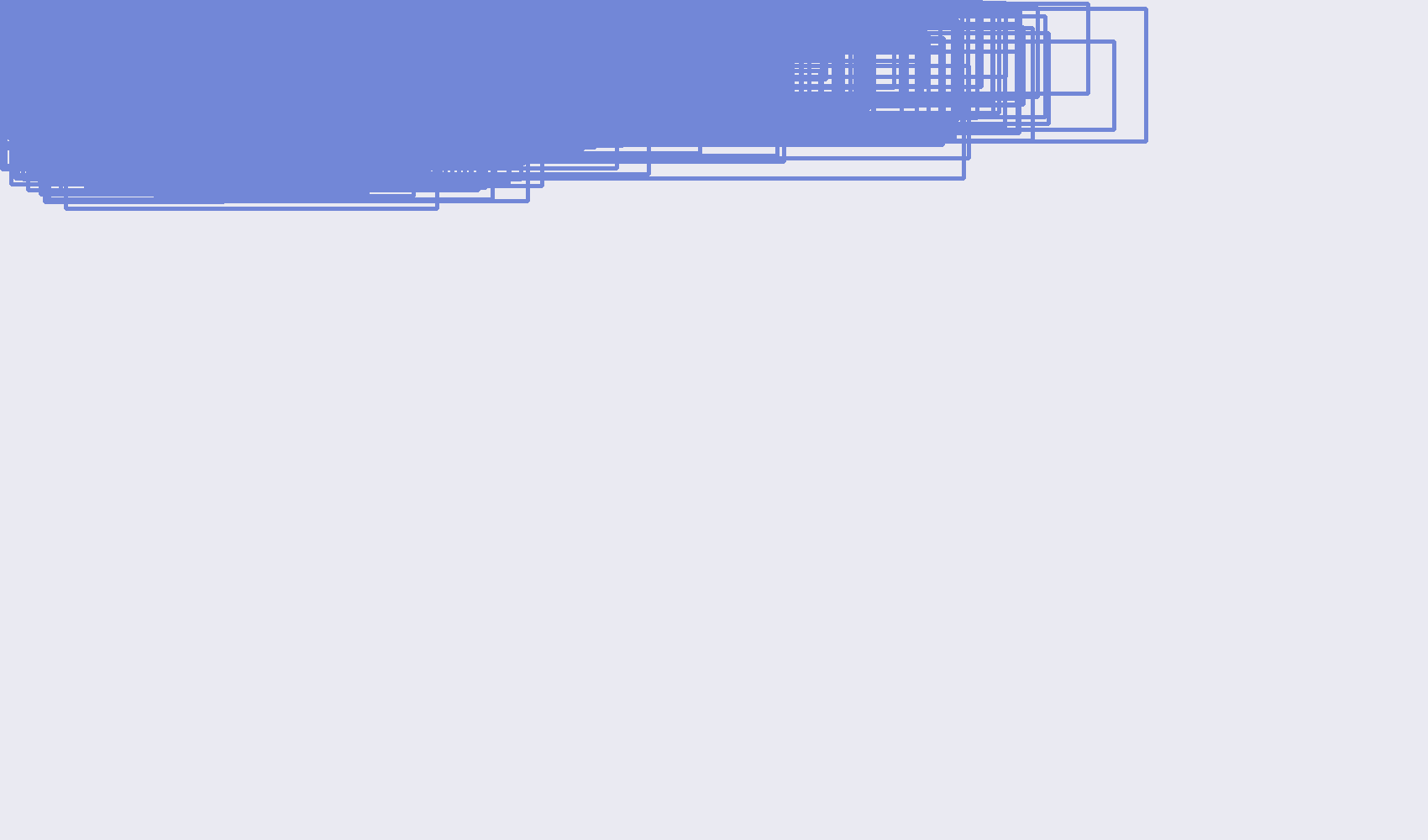}
    \caption{Bounding boxes of all instances to which the rule \textit{largest bounding box in the upper left corner} was applied to.}
    \label{fig:yolo_all_bb_after}
\end{figure}

\section{Training details}
\label{a:train_details}

We used the defaults from \texttt{transformers} (4.26.1), if not reported otherwise.

\subsection{Standard Training}

\begin{table}[H]
    \centering
    \begin{tabular}{l|l}
        Parameter & Value \\ \hline \hline
        Seed & 42 \\
        Optimizer & AdamW \\
        Epochs & 5 \\
        Decoder & GPT-2 \\
        Encoder & \{BEIT, Swin, ViT\} \\
        Batch Size (Train \& Test) & 64 \\
    \end{tabular}
    \caption{Parameters for the standard training.}
    \label{tab:params_tr}
\end{table}

\subsection{Training with Augmentation}
\begin{table}[H]
    \centering
    \begin{tabular}{l|l}
        Parameter & Value \\ \hline \hline
        Seed & 42 \\
        Optimizer & AdamW \\
        Epochs & \{5, 20\} \\
        Decoder & GPT-2 \\
        Encoder & \{BEIT, Swin, ViT\} \\
        Batch Size (Train \& Test) & 64 \\
    \end{tabular}
    \caption{Parameters for training with augmentation.}
    \label{tab:params_tr_aug}
\end{table}

\subsection{Natural Language Generation}
\begin{table}[H]
    \centering
    \begin{tabular}{l|l}
        Parameter & Value \\ \hline \hline
        Max Length & 32 \\
        Early Stopping & True \\
        No Repeat Ngram Size & 3 \\
        Length Penalty & 2.0 \\
        Number of Beams & 4 \\
        
    \end{tabular}
    \caption{Parameters for natural language generation.}
    \label{tab:params_nlg}
\end{table}

\subsection{Decoder Pre-Training}
\begin{table}[H]
    \centering
    \begin{tabular}{l|l}
        Parameter & Value \\ \hline \hline
        Seed & 42 \\
        Epochs & 10 \\
        Batch Size (Train \& Test) & 192 \\
    \end{tabular}
    \caption{Parameters for pre-training of the decoder.}
    \label{tab:params_ptd}
\end{table}



\begin{acronym}
 \acro{BEiT}{Bidirectional Encoder representation for Image Transformers}
 \acro{CER}{Character Error Rate}
 \acro{CNN}{Convolutional Neural Network}
 \acro{CTC}{Connectionist Temporal Classification}
 \acro{GCV}{Gloogle Cloud Vision}
 \acro{HTR}{Handwritten Text Recognition}
 \acro{MLW}{Medieval Latin Dictionary}
 \acro{OCR}{Optical Character Recognition}
 \acro{OFA}{One For All}
 \acro{RNN}{Recurrent Neural Network}
 \acro{SOTA}{state-of-the-art}
 \acro{Swin}{Shifted Window Transformer}
 \acro{ViT}{Vision Transformer}
 \acro{YOLO}{You Only Look Once}
\end{acronym}

\end{document}